\pdfoutput=1
\documentclass[%
 reprint,
 amsmath,amssymb,
 aps,
]{revtex4-2}
\usepackage{ccaption}
\usepackage{hyperref}
\hypersetup{
    colorlinks=true,
    linkcolor=blue,
    filecolor=magenta,      
    urlcolor=cyan,
    }

\usepackage{amsmath}
\usepackage{graphicx}
\usepackage{dcolumn}
\usepackage{bm}

\begin{document}

\preprint{APS/123-QED}

\title{A neural net architecture based on principles of neural plasticity and development evolves to effectively catch prey in a simulated environment}

\author{Addison Wood}
\email{woodadi@msu.edu}
 \affiliation{Neuroscience Program, Michigan State University}
\author{Jory Schossau}
\email{jory@msu.edu}
 \affiliation{BEACON Center, Michigan State University}
\author{Nick Sabaj}
\author{Richard Liu}
\author{Mark Reimers}
\email{reimersm@msu.edu}
 \affiliation{Neuroscience Program, Michigan State University}


\date{\today}

\begin{abstract}
A profound challenge for A-Life is to construct agents whose behavior is `life-like' in a deep way. We propose an architecture and approach to constructing networks driving artificial agents, using processes analogous to the processes that construct and sculpt the brains of animals. Furthermore, the instantiation of action is dynamic: the whole network responds in real-time to sensory inputs to activate effectors, rather than computing a representation of the optimal behavior and sending off an encoded representation to effector controllers. There are many parameters and we use an evolutionary algorithm to select them, in the context of a specific prey-capture task. We think this architecture may be useful for controlling small autonomous robots or drones, because it allows for a rapid response to changes in sensor inputs. 
\end{abstract}

\maketitle


\section{\label{sec:intro}introduction}

One of the most profound challenges of research into artificial life (A-Life) is to construct agents whose behavior is generated in ways deeply analogous to the way an animal's brain drives its motions. Many researchers argue for using neural networks to control behavior \cite{richter2016scalability, Amer_Samy_Shaker_ElHelw_2021, Hole_Ahmad_2021, Hole_Ahmad_2021} and some favor an approach that builds an ersatz nervous system, whose operations mimic aspects of an animal's nervous system \cite{Stanley_Clune_Lehman_Miikkulainen_2019, Duarte_Gomes_Oliveira_Christensen_2018, Sheneman_Schossau_Hintze_2019}. 

However, there is no consensus about how closely construction must hew to biology Furthermore, for large neural networks, the number of possible network configurations is so large that it is difficult to search for network constructions that drive adaptive or effective behavior. Evolutionary algorithms have a solid track record for solving problems with a large search space \cite{Goodman_Rothwell_Averill_2011}.  We introduce here an evolutionary approach to constructing a neural network controller for a small animat in a 2-dimensional world. 

Our approach is distinctive in several ways. First, behavior will emerge through dynamic equilibria of the network and continuous feedback from the surroundings carried by a rich array of simple sensors, rather than having the network compute a representation of behavior, which is then implemented by mechanical systems. The network architecture will be recurrent feedback loops, analogous to loops between cortex and subcortical structures, such as thalamus and basal ganglia. Second, neurons will compete and oppose the actions of other neurons, so that the dynamic equilibrium may be rapidly  altered as circumstances change. Third, each instance of a controller will be constructed by methods modeled on developmental processes in animal brains, as detailed below. To our knowledge, this is the most life-like approach yet taken to constructing A-life simulations.

\section{\label{sec:methods}Methods}
\subsection{\label{sec:construction}Animat Construction}

Our system builds neural networks in three stages, modeled on neurobiology: 
\paragraph{Node definition} Nodes are assigned random pseudo-locations in a 2D, 3D or 4D virtual space. Parameters define the probabilities of specific node classes at various locations within the virtual space, and within a class, the specific signals employed for further stages.
\paragraph{Coarse initial wiring} Projections are assigned with synaptic weights determined by the match between paired transmitter and receiver signals, in a manner intended to be analogous to how axons find distant targets through molecular interactions in the developing brain \cite{Hassan_Hiesinger_2015, Rubenstein_Rakic_2013}. In our prototype system the signals are determined by gradients in 2D virtual space determined by parameters.
\paragraph{Fine mapping} Cells sort out connections through periods of spontaneous activity and Hebbian plasticity (``cells that fire together wire together'') \cite{Tiriac_Feller_2019, Kerschensteiner_2014}.
\paragraph{Learning} Experiences induce changes in synapses, in a Hebbian manner, contingent upon attaining food, mediated by neuromodulator (NM) signals. The intrinsically unstable Hebbian process is further stabilized in two ways: \textbf{i)} saturation of the plasticity signal, and \textbf{ii)} longer-term homeostatic mechanisms, both modeled on neurobiology \cite{escobar2007long, Turrigiano_2012}. The processes of synapse generation at each stage are determined by a small number of parameters. During the subsequent learning stage, synaptic weights are periodically adjusted according to a pseudo-Hebbian learning rule, modulated by reward signals, whose parameters are set by genes.

\subsection{\label{sec:task}Evaluation Task}

The simulated world is a two dimensional plane, populated by two species of prey items, both of which move according to a Brownian Motion with a strong drift away from the animat, when the animat is within striking distance; their top speed is slightly less than the animat's top speed. One prey species, comprising 1/3 of all prey, has more calories than the other; which species is to have more calories is chosen randomly in each generation, so that the genome cannot evolve to encode that value, but the animat must learn it in each generation. The fitness function for the animat is how many calories are captured per lifetime.

\subsection{\label{sec:simulation}Simulation}

The evolutionary algorithm was implemented by highly efficient special purpose open source software written in C++ that has been developed at MSU: MABE (Modular Agent Based Evolution framework) \cite{Bohm_Hintze_2017}. MABE allows researchers to combine different forms of cognitive substrates, tasks, selection methods and data collection methods to quickly develop digital neuroevolution experiments. MABE is ideally suited for this work since it allows parts of an experiment to be swapped without affecting other parts, which allows us to easily compare the behavior of the animat with other more popular methods of computational neuroevolution.
Since the parameters that define the construction process and learning characteristics are continuous or integer-valued (with a large range of values), our mutation operator works by perturbing numerical values, rather than swapping symbols in a string. The size of perturbation is initially set to a plausible value about 10\% of the range of values considered reasonable. See Appendix item \MakeUppercase{\romannumeral 1} for the initial values of mutation and range used for the simulations described here.

\section{Results}
\begin{figure*}
\centering
\includegraphics[width=0.90\textwidth]{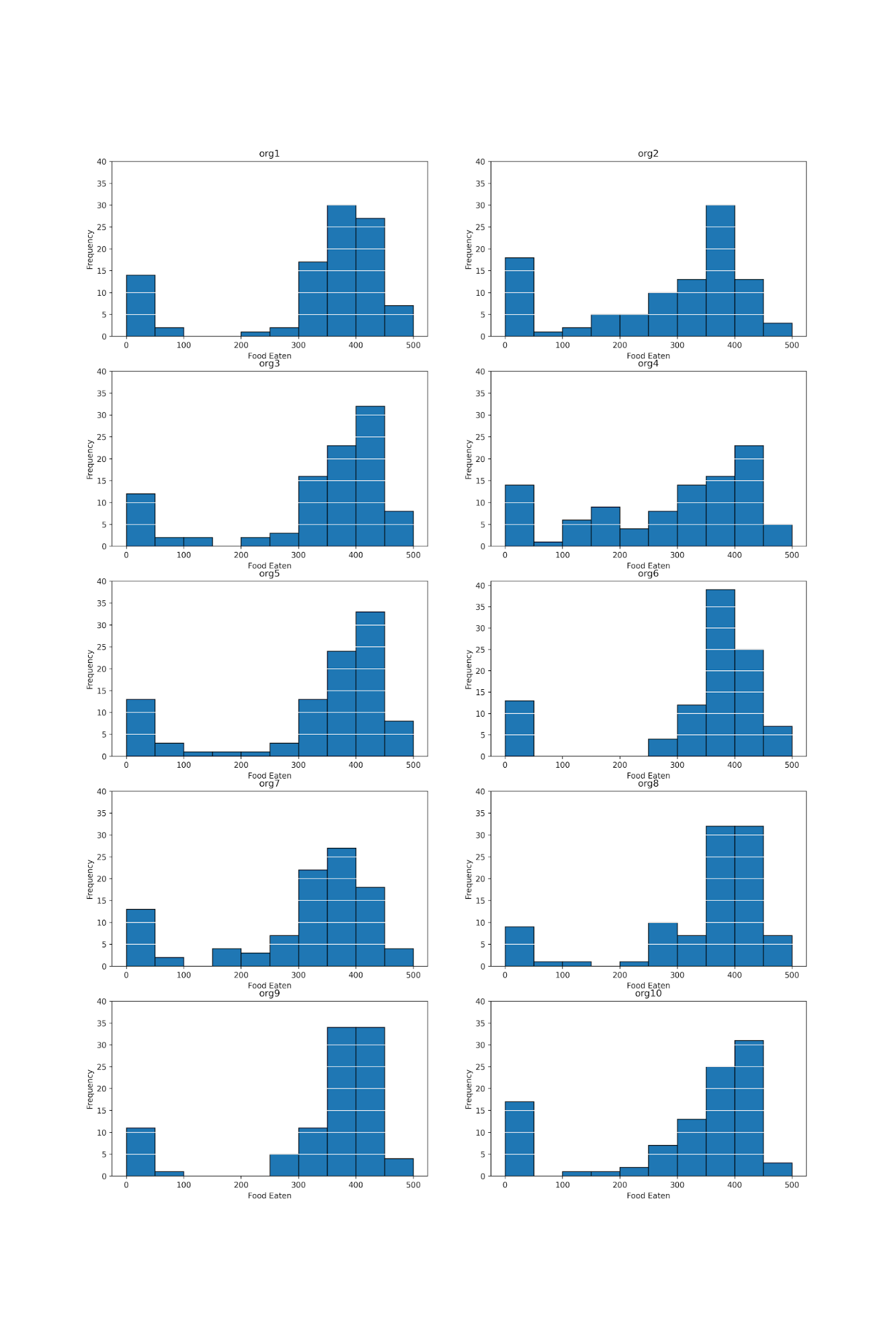}
\caption{\label{fig:fooddist10}
Distribution of foods eaten by 10 successful agents, each evaluated in 100 different randomly-generated worlds. Full folder of videos, plots, and histograms for all 10 runs: \href{https://drive.google.com/drive/folders/1GjFWGk9yo5Ls56YX44YtIAW68TWAu70d?usp=sharing}{ \texttt{Animat Videos \& Figures} }
}
\end{figure*}

In our simulations, agents started by capturing an average of 2 or 3 food items in generation 1 and usually attained an average of ~25 food items per 5 minutes simulated lifetime after 50 or 100 generations at which point their performance stabilized until several thousands of generations, at which point there typically was a rapid increase in fitness, which then plateaued at several hundred food items captured, while showing a strong preference for the high-calorie prey. After the second larger jump in most runs, in the stable population regime, agents ate an average of 282 prey items, with a standard deviation of 151. After performance of the top 10
One issue for assessing performance, is how much the measure reflects the genome and how much it reflects (random) starting conditions. To test this we extracted 10 successful agents within the stable population regime, and then placed them in 100 random worlds, in which the starting location of the prey is randomly generated, we see that the distribution of food eaten for each agent throughout the random worlds (\ref{fig:fooddist10}) begins to look similar to the histogram of food eaten for all agents (successful or not) in the stable population regime. Looking at the histograms visualizing these count numbers, it's easy to spot the pattern: the lackluster agents consume next to nothing, there is a relatively large gap in-between, then there is a larger group of well-performing agents arranged in a (left-skewed) normal distribution.
The most successful agents were exceptional about identifying and chasing prey. These agents seemed to set and follow a trajectory for the nearest cluster of prey items, then after momentum carried the agent out of the group, they could loop back around or find the new nearest cluster of prey, which is an impressive behavior, as the agent moves quite quickly. The agent can traverse the entire range encompassing all 50 prey in 1 second of simulated time. This is in contrast to the behavior of agents before the jump, whose movement was characterized by a large, consistent looping pattern - a relatively successful strategy, but largely inefficient, which we termed ``filter feeding.''
Evolutionary theorists often ask how inevitable a specific evolutionary outcome is. Although so far the large majority of our simulations have found a stable high-performing population, the distributions of construction parameters (`genes') are very different from each other, any pair overlapping less than 1\%. This indicates that even in this simple 22-parameter system a great many distinct possible optima, with comparable performance, may be found.

\begin{figure*}
\includegraphics[width=\textwidth]{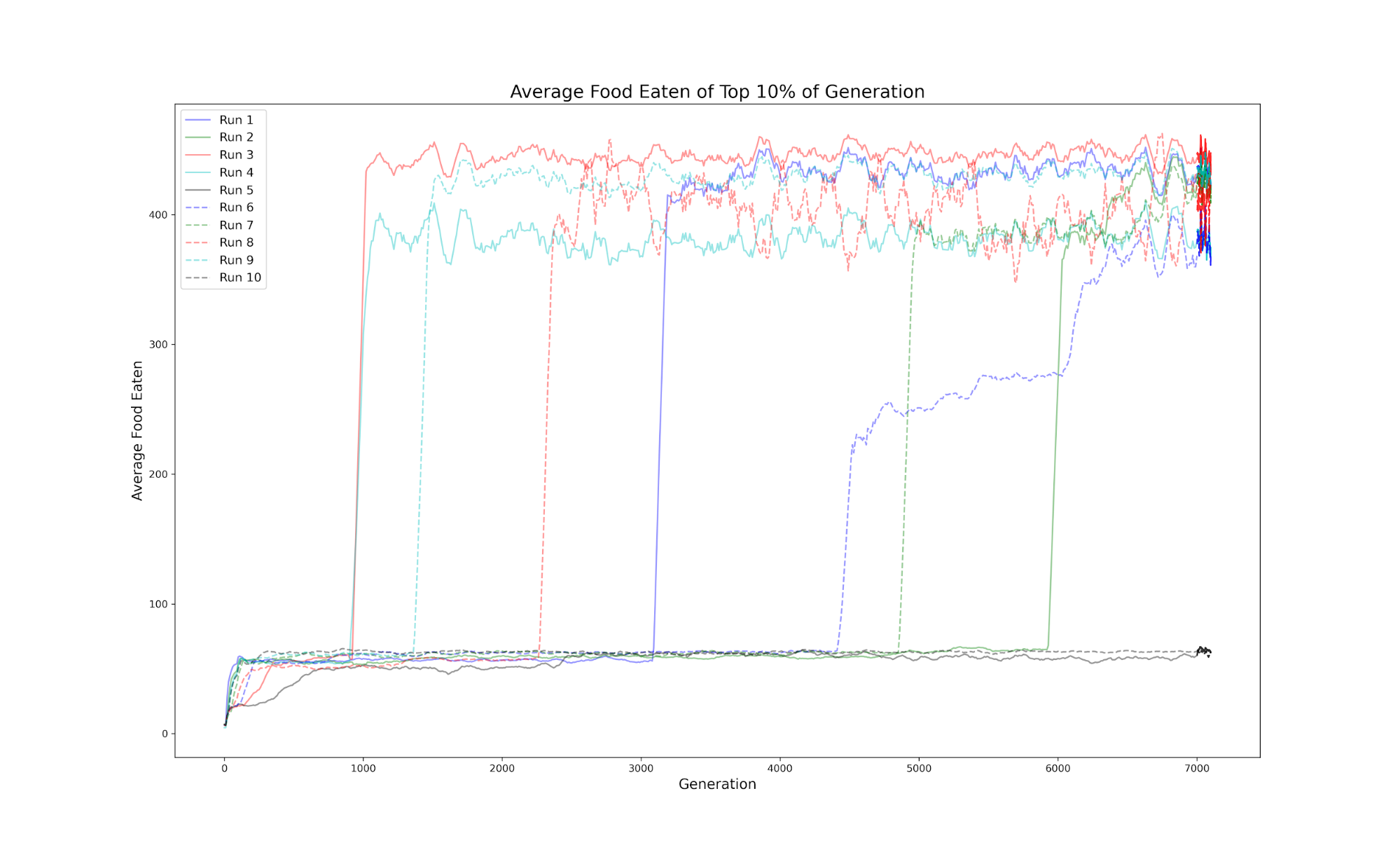}
\caption{\label{fig:scoreevolution}Typical example of fitness (foods eaten) among top 10\% of animats as a function of generation. The solid section at the end represents the `stable population' period: where we narrowed the effect of the mutation operator for genes that had a narrow range of values for successful agents, in order to keep most offspring within plausible range of viability.}
\end{figure*}
\begin{acknowledgments}
\end{acknowledgments}

\begin{figure*}
\includegraphics[width=\textwidth]{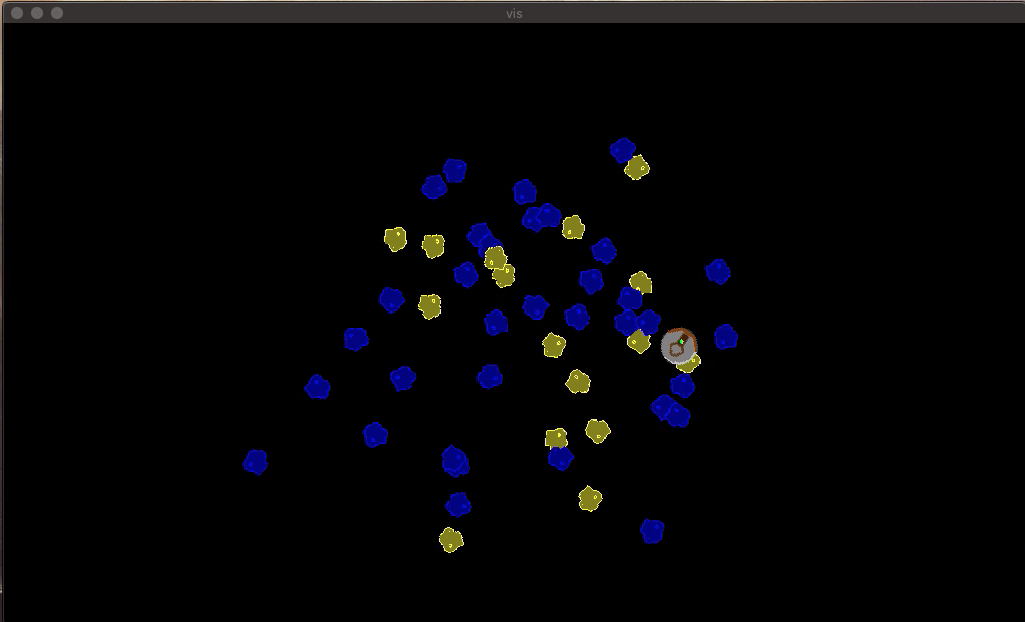}
\caption{\label{fig:visualization} Visualization of the animat (large round beige object) and prey - link to movie: \href{https://drive.google.com/file/d/1_exJlo8ZFmy4oKJH0wi3EoDGIJp2P9Ig/view?usp=sharing}{ \texttt{animat\_video\_01.mov} } The Roomba-like object is the animat, where the black area in the top-right of the agent represents its front. The blue shapes are low-calorie prey, while the yellow shapes are  high-calorie prey. It is important to note when viewing the full video that it is $0.3\times$ the normal speed, so that the behavior of the agent can be more easily tracked.}
\end{figure*}

\begin{figure*}
\includegraphics[width=\textwidth]{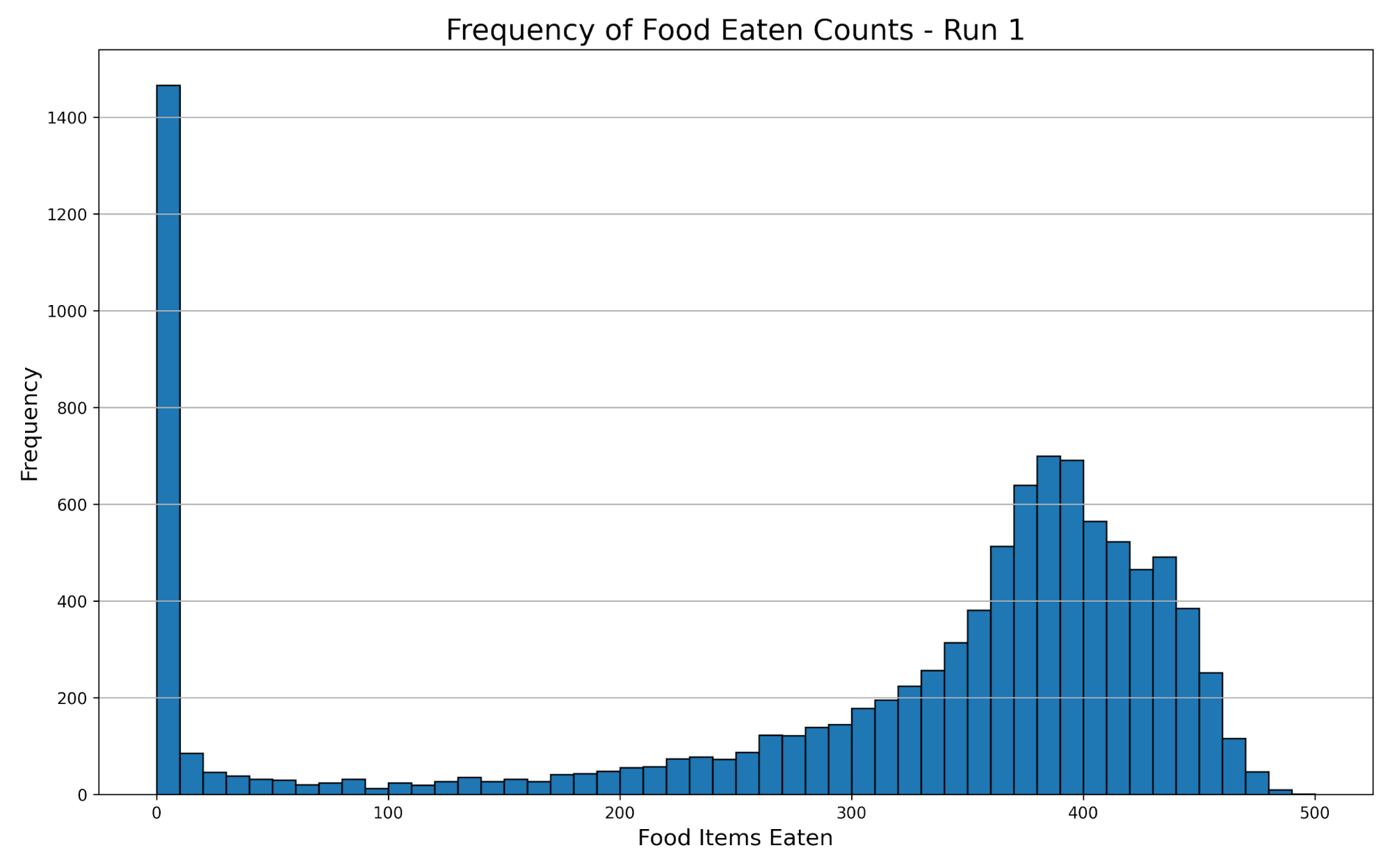}
\caption{\label{fig:fooddist1} Distribution of foods eaten during lifetime (5 min) by animat agents during 100 generations of the stable population ($N=100$ per generation)}
\end{figure*}

\clearpage
\nocite{*}
\bibliography{ms}

\providecommand{\noopsort}[1]{}\providecommand{\singleletter}[1]{#1}%
\begin{thebibliography}{15}%
\makeatletter
\providecommand \@ifxundefined [1]{%
 \@ifx{#1\undefined}
}%
\providecommand \@ifnum [1]{%
 \ifnum #1\expandafter \@firstoftwo
 \else \expandafter \@secondoftwo
 \fi
}%
\providecommand \@ifx [1]{%
 \ifx #1\expandafter \@firstoftwo
 \else \expandafter \@secondoftwo
 \fi
}%
\providecommand \natexlab [1]{#1}%
\providecommand \enquote  [1]{``#1''}%
\providecommand \bibnamefont  [1]{#1}%
\providecommand \bibfnamefont [1]{#1}%
\providecommand \citenamefont [1]{#1}%
\providecommand \href@noop [0]{\@secondoftwo}%
\providecommand \href [0]{\begingroup \@sanitize@url \@href}%
\providecommand \@href[1]{\@@startlink{#1}\@@href}%
\providecommand \@@href[1]{\endgroup#1\@@endlink}%
\providecommand \@sanitize@url [0]{\catcode `\\12\catcode `\$12\catcode
  `\&12\catcode `\#12\catcode `\^12\catcode `\_12\catcode `\%12\relax}%
\providecommand \@@startlink[1]{}%
\providecommand \@@endlink[0]{}%
\providecommand \url  [0]{\begingroup\@sanitize@url \@url }%
\providecommand \@url [1]{\endgroup\@href {#1}{\urlprefix }}%
\providecommand \urlprefix  [0]{URL }%
\providecommand \Eprint [0]{\href }%
\providecommand \doibase [0]{https://doi.org/}%
\providecommand \selectlanguage [0]{\@gobble}%
\providecommand \bibinfo  [0]{\@secondoftwo}%
\providecommand \bibfield  [0]{\@secondoftwo}%
\providecommand \translation [1]{[#1]}%
\providecommand \BibitemOpen [0]{}%
\providecommand \bibitemStop [0]{}%
\providecommand \bibitemNoStop [0]{.\EOS\space}%
\providecommand \EOS [0]{\spacefactor3000\relax}%
\providecommand \BibitemShut  [1]{\csname bibitem#1\endcsname}%
\let\auto@bib@innerbib\@empty
\bibitem [{\citenamefont {Richter}\ \emph {et~al.}(2016)\citenamefont
  {Richter}, \citenamefont {Jentzsch}, \citenamefont {Hostettler},
  \citenamefont {Garrido}, \citenamefont {Ros}, \citenamefont {Knoll},
  \citenamefont {R{\"o}hrbein}, \citenamefont {van~der Smagt},\ and\
  \citenamefont {Conradt}}]{richter2016scalability}%
  \BibitemOpen
  \bibfield  {author} {\bibinfo {author} {\bibfnamefont {C.}~\bibnamefont
  {Richter}}, \bibinfo {author} {\bibfnamefont {S.}~\bibnamefont {Jentzsch}},
  \bibinfo {author} {\bibfnamefont {R.}~\bibnamefont {Hostettler}}, \bibinfo
  {author} {\bibfnamefont {J.~A.}\ \bibnamefont {Garrido}}, \bibinfo {author}
  {\bibfnamefont {E.}~\bibnamefont {Ros}}, \bibinfo {author} {\bibfnamefont
  {A.~C.}\ \bibnamefont {Knoll}}, \bibinfo {author} {\bibfnamefont
  {F.}~\bibnamefont {R{\"o}hrbein}}, \bibinfo {author} {\bibfnamefont
  {P.}~\bibnamefont {van~der Smagt}},\ and\ \bibinfo {author} {\bibfnamefont
  {J.}~\bibnamefont {Conradt}},\ }\bibfield  {title} {\bibinfo {title}
  {Scalability in neural control of musculoskeletal robots},\ }\href@noop {}
  {\bibfield  {journal} {\bibinfo  {journal} {arXiv preprint arXiv:1601.04862}\
  } (\bibinfo {year} {2016})}\BibitemShut {NoStop}%
\bibitem [{\citenamefont {Amer}\ \emph {et~al.}(2021)\citenamefont {Amer},
  \citenamefont {Samy}, \citenamefont {Shaker},\ and\ \citenamefont
  {ElHelw}}]{Amer_Samy_Shaker_ElHelw_2021}%
  \BibitemOpen
  \bibfield  {author} {\bibinfo {author} {\bibfnamefont {K.}~\bibnamefont
  {Amer}}, \bibinfo {author} {\bibfnamefont {M.}~\bibnamefont {Samy}}, \bibinfo
  {author} {\bibfnamefont {M.}~\bibnamefont {Shaker}},\ and\ \bibinfo {author}
  {\bibfnamefont {M.}~\bibnamefont {ElHelw}},\ }\bibfield  {title} {\bibinfo
  {title} {Deep convolutional neural network based autonomous drone
  navigation},\ }in\ \href {https://doi.org/10.1117/12.2587105} {\emph
  {\bibinfo {booktitle} {Thirteenth International Conference on Machine
  Vision}}},\ Vol.\ \bibinfo {volume} {11605}\ (\bibinfo  {publisher} {SPIE},\
  \bibinfo {year} {2021})\ p.\ \bibinfo {pages} {16–24}\BibitemShut {NoStop}%
\bibitem [{\citenamefont {Hole}\ and\ \citenamefont
  {Ahmad}(2021)}]{Hole_Ahmad_2021}%
  \BibitemOpen
  \bibfield  {author} {\bibinfo {author} {\bibfnamefont {K.~J.}\ \bibnamefont
  {Hole}}\ and\ \bibinfo {author} {\bibfnamefont {S.}~\bibnamefont {Ahmad}},\
  }\bibfield  {title} {\bibinfo {title} {A thousand brains: toward biologically
  constrained ai},\ }\href {https://doi.org/10.1007/s42452-021-04715-0}
  {\bibfield  {journal} {\bibinfo  {journal} {SN Applied Sciences}\ }\textbf
  {\bibinfo {volume} {3}},\ \bibinfo {pages} {743} (\bibinfo {year}
  {2021})}\BibitemShut {NoStop}%
\bibitem [{\citenamefont {Stanley}\ \emph {et~al.}(2019)\citenamefont
  {Stanley}, \citenamefont {Clune}, \citenamefont {Lehman},\ and\ \citenamefont
  {Miikkulainen}}]{Stanley_Clune_Lehman_Miikkulainen_2019}%
  \BibitemOpen
  \bibfield  {author} {\bibinfo {author} {\bibfnamefont {K.~O.}\ \bibnamefont
  {Stanley}}, \bibinfo {author} {\bibfnamefont {J.}~\bibnamefont {Clune}},
  \bibinfo {author} {\bibfnamefont {J.}~\bibnamefont {Lehman}},\ and\ \bibinfo
  {author} {\bibfnamefont {R.}~\bibnamefont {Miikkulainen}},\ }\bibfield
  {title} {\bibinfo {title} {Designing neural networks through
  neuroevolution},\ }\href {https://doi.org/10.1038/s42256-018-0006-z}
  {\bibfield  {journal} {\bibinfo  {journal} {Nature Machine Intelligence}\
  }\textbf {\bibinfo {volume} {1}},\ \bibinfo {pages} {24–35} (\bibinfo
  {year} {2019})}\BibitemShut {NoStop}%
\bibitem [{\citenamefont {Duarte}\ \emph {et~al.}(2018)\citenamefont {Duarte},
  \citenamefont {Gomes}, \citenamefont {Oliveira},\ and\ \citenamefont
  {Christensen}}]{Duarte_Gomes_Oliveira_Christensen_2018}%
  \BibitemOpen
  \bibfield  {author} {\bibinfo {author} {\bibfnamefont {M.}~\bibnamefont
  {Duarte}}, \bibinfo {author} {\bibfnamefont {J.}~\bibnamefont {Gomes}},
  \bibinfo {author} {\bibfnamefont {S.~M.}\ \bibnamefont {Oliveira}},\ and\
  \bibinfo {author} {\bibfnamefont {A.~L.}\ \bibnamefont {Christensen}},\
  }\bibfield  {title} {\bibinfo {title} {Evolution of repertoire-based control
  for robots with complex locomotor systems},\ }\href
  {https://doi.org/10.1109/TEVC.2017.2722101} {\bibfield  {journal} {\bibinfo
  {journal} {IEEE Transactions on Evolutionary Computation}\ }\textbf {\bibinfo
  {volume} {22}},\ \bibinfo {pages} {314–328} (\bibinfo {year}
  {2018})}\BibitemShut {NoStop}%
\bibitem [{\citenamefont {Sheneman}\ \emph {et~al.}(2019)\citenamefont
  {Sheneman}, \citenamefont {Schossau},\ and\ \citenamefont
  {Hintze}}]{Sheneman_Schossau_Hintze_2019}%
  \BibitemOpen
  \bibfield  {author} {\bibinfo {author} {\bibfnamefont {L.}~\bibnamefont
  {Sheneman}}, \bibinfo {author} {\bibfnamefont {J.}~\bibnamefont {Schossau}},\
  and\ \bibinfo {author} {\bibfnamefont {A.}~\bibnamefont {Hintze}},\
  }\bibfield  {title} {\bibinfo {title} {The evolution of neuroplasticity and
  the effect on integrated information},\ }\href
  {https://doi.org/10.3390/e21050524} {\bibfield  {journal} {\bibinfo
  {journal} {Entropy}\ }\textbf {\bibinfo {volume} {21}},\ \bibinfo {pages}
  {524} (\bibinfo {year} {2019})}\BibitemShut {NoStop}%
\bibitem [{\citenamefont {Goodman}\ \emph {et~al.}(2011)\citenamefont
  {Goodman}, \citenamefont {Rothwell},\ and\ \citenamefont
  {Averill}}]{Goodman_Rothwell_Averill_2011}%
  \BibitemOpen
  \bibfield  {author} {\bibinfo {author} {\bibfnamefont {E.~D.}\ \bibnamefont
  {Goodman}}, \bibinfo {author} {\bibfnamefont {E.~J.}\ \bibnamefont
  {Rothwell}},\ and\ \bibinfo {author} {\bibfnamefont {R.~C.}\ \bibnamefont
  {Averill}},\ }\bibfield  {title} {\bibinfo {title} {Using concepts from
  biology to improve problem-solving methods},\ }in\ \href
  {https://doi.org/10.1117/12.889070} {\emph {\bibinfo {booktitle}
  {Evolutionary and Bio-Inspired Computation: Theory and Applications V}}},\
  Vol.\ \bibinfo {volume} {8059}\ (\bibinfo  {publisher} {SPIE},\ \bibinfo
  {year} {2011})\ p.\ \bibinfo {pages} {9–24}\BibitemShut {NoStop}%
\bibitem [{\citenamefont {Hassan}\ and\ \citenamefont
  {Hiesinger}(2015)}]{Hassan_Hiesinger_2015}%
  \BibitemOpen
  \bibfield  {author} {\bibinfo {author} {\bibfnamefont {B.~A.}\ \bibnamefont
  {Hassan}}\ and\ \bibinfo {author} {\bibfnamefont {P.~R.}\ \bibnamefont
  {Hiesinger}},\ }\bibfield  {title} {\bibinfo {title} {Beyond molecular codes:
  Simple rules to wire complex brains},\ }\href
  {https://doi.org/10.1016/j.cell.2015.09.031} {\bibfield  {journal} {\bibinfo
  {journal} {Cell}\ }\textbf {\bibinfo {volume} {163}},\ \bibinfo {pages}
  {285–291} (\bibinfo {year} {2015})}\BibitemShut {NoStop}%
\bibitem [{\citenamefont {Rubenstein}\ and\ \citenamefont
  {Rakic}(2013)}]{Rubenstein_Rakic_2013}%
  \BibitemOpen
  \bibfield  {author} {\bibinfo {author} {\bibfnamefont {J.}~\bibnamefont
  {Rubenstein}}\ and\ \bibinfo {author} {\bibfnamefont {P.}~\bibnamefont
  {Rakic}},\ }\href@noop {} {\emph {\bibinfo {title} {Cellular migration and
  formation of neuronal connections: comprehensive developmental
  neuroscience}}},\ Vol.~\bibinfo {volume} {2}\ (\bibinfo  {publisher}
  {Academic Press},\ \bibinfo {year} {2013})\BibitemShut {NoStop}%
\bibitem [{\citenamefont {Tiriac}\ and\ \citenamefont
  {Feller}(2019)}]{Tiriac_Feller_2019}%
  \BibitemOpen
  \bibfield  {author} {\bibinfo {author} {\bibfnamefont {A.}~\bibnamefont
  {Tiriac}}\ and\ \bibinfo {author} {\bibfnamefont {M.~B.}\ \bibnamefont
  {Feller}},\ }\bibfield  {title} {\bibinfo {title} {Embryonic neural activity
  wires the brain},\ }\href@noop {} {\bibfield  {journal} {\bibinfo  {journal}
  {Science}\ }\textbf {\bibinfo {volume} {364}},\ \bibinfo {pages} {933–934}
  (\bibinfo {year} {2019})}\BibitemShut {NoStop}%
\bibitem [{\citenamefont {Kerschensteiner}(2014)}]{Kerschensteiner_2014}%
  \BibitemOpen
  \bibfield  {author} {\bibinfo {author} {\bibfnamefont {D.}~\bibnamefont
  {Kerschensteiner}},\ }\bibfield  {title} {\bibinfo {title} {Spontaneous
  network activity and synaptic development},\ }\href
  {https://doi.org/10.1177/1073858413510044} {\bibfield  {journal} {\bibinfo
  {journal} {The Neuroscientist: A Review Journal Bringing Neurobiology,
  Neurology and Psychiatry}\ }\textbf {\bibinfo {volume} {20}},\ \bibinfo
  {pages} {272–290} (\bibinfo {year} {2014})}\BibitemShut {NoStop}%
\bibitem [{\citenamefont {Escobar}\ and\ \citenamefont
  {Derrick}(2007)}]{escobar2007long}%
  \BibitemOpen
  \bibfield  {author} {\bibinfo {author} {\bibfnamefont {M.~L.}\ \bibnamefont
  {Escobar}}\ and\ \bibinfo {author} {\bibfnamefont {B.}~\bibnamefont
  {Derrick}},\ }\bibfield  {title} {\bibinfo {title} {Long-term potentiation
  and depression as putative mechanisms for memory formation},\ }\href@noop {}
  {\bibfield  {journal} {\bibinfo  {journal} {Neural Plasticity and Memory:
  From Genes to Brain Imaging}\ ,\ \bibinfo {pages} {15}} (\bibinfo {year}
  {2007})}\BibitemShut {NoStop}%
\bibitem [{\citenamefont {Turrigiano}(2012)}]{Turrigiano_2012}%
  \BibitemOpen
  \bibfield  {author} {\bibinfo {author} {\bibfnamefont {G.}~\bibnamefont
  {Turrigiano}},\ }\bibfield  {title} {\bibinfo {title} {Homeostatic synaptic
  plasticity: local and global mechanisms for stabilizing neuronal function},\
  }\href@noop {} {\bibfield  {journal} {\bibinfo  {journal} {Cold Spring Harbor
  perspectives in biology}\ }\textbf {\bibinfo {volume} {4}},\ \bibinfo {pages}
  {a005736} (\bibinfo {year} {2012})}\BibitemShut {NoStop}%
\bibitem [{\citenamefont {Bohm}\ and\ \citenamefont
  {Hintze}(2017)}]{Bohm_Hintze_2017}%
  \BibitemOpen
  \bibfield  {author} {\bibinfo {author} {\bibfnamefont {C.}~\bibnamefont
  {Bohm}}\ and\ \bibinfo {author} {\bibfnamefont {A.}~\bibnamefont {Hintze}},\
  }\bibfield  {title} {\bibinfo {title} {Mabe (modular agent based evolver): A
  framework for digital evolution research},\ }in\ \href@noop {} {\emph
  {\bibinfo {booktitle} {ECAL 2017, the Fourteenth European Conference on
  Artificial Life}}}\ (\bibinfo  {publisher} {MIT Press},\ \bibinfo {year}
  {2017})\ p.\ \bibinfo {pages} {76–83}\BibitemShut {NoStop}%
\bibitem [{\citenamefont {Amunts}\ \emph {et~al.}(2016)\citenamefont {Amunts},
  \citenamefont {Ebell}, \citenamefont {Muller}, \citenamefont {Telefont},
  \citenamefont {Knoll},\ and\ \citenamefont
  {Lippert}}]{Amunts_Ebell_Muller_Telefont_Knoll_Lippert_2016}%
  \BibitemOpen
  \bibfield  {author} {\bibinfo {author} {\bibfnamefont {K.}~\bibnamefont
  {Amunts}}, \bibinfo {author} {\bibfnamefont {C.}~\bibnamefont {Ebell}},
  \bibinfo {author} {\bibfnamefont {J.}~\bibnamefont {Muller}}, \bibinfo
  {author} {\bibfnamefont {M.}~\bibnamefont {Telefont}}, \bibinfo {author}
  {\bibfnamefont {A.}~\bibnamefont {Knoll}},\ and\ \bibinfo {author}
  {\bibfnamefont {T.}~\bibnamefont {Lippert}},\ }\bibfield  {title} {\bibinfo
  {title} {The human brain project: Creating a european research infrastructure
  to decode the human brain},\ }\href
  {https://doi.org/10.1016/j.neuron.2016.10.046} {\bibfield  {journal}
  {\bibinfo  {journal} {Neuron}\ }\textbf {\bibinfo {volume} {92}},\ \bibinfo
  {pages} {574–581} (\bibinfo {year} {2016})}\BibitemShut {NoStop}%
\end{thebibliography}%


\end{document}


\appendix
\section{Genomic Properties}

\begin{table}
\begin{minipage}{\textwidth}
\vspace{4ex}
\centering
\caption{\label{tab:parameters}Genome of animat parameters and the mutation scheme stochastically applied.}
\begin{ruledtabular}
\begin{tabular}{lccc}
Gene&Initial mutation standard deviation&Lower Bound&Upper Bound\\ \hline

$\text{Ligand}_1$         & 0.5        & 0     &   12 \\
$\text{Ligand}_2$         & 0.5        & 0     &   12 \\
$\text{Ligand}_3$         & 0.5        & 0     &   12 \\
$\text{Ligand}_4$         & 0.5        & 0     &   12 \\
$\text{Ligand}_5$         & 0.5        & 0     &   12 \\
$\text{Receptor}_1$       & 0.5        & 0     &   12 \\
$\text{Receptor}_2$       & 0.5        & 0     &   12 \\
$\text{Receptor}_3$       & 0.5        & 0     &   12 \\
$\text{Receptor}_4$       & 0.5        & 0     &   12 \\
$\text{Receptor}_5$       & 0.5        & 0     &   12 \\
$\text{sensory}_{\mathrm{adapt-min-scale-s0}}$    & 0.1        & 0     &    1 \\
$\text{sensory}_{\mathrm{adapt-delta-scale-a1}}$    & 0.5        & 0     &    1 \\
$\text{sensory}_{\mathrm{adapt-delta-threshold-a2}}$    & 0.5        & 0     &   20 \\
$\text{homeostatic}$      & 0.1        & 0     &    1 \\
$\text{decay}_{\mathrm{proximity}}$ & 100        & 0     & 2000 \\
$\text{decay}_{\mathrm{eating}}$   & 100        & 0     & 2000 \\
$\text{decay}_{\mathrm{hunger}}$   & 100        & 0     & 2000 \\

\end{tabular}
\end{ruledtabular}
\end{minipage}
\end{table}